\PassOptionsToPackage{dvipsnames,table,xcdraw}{xcolor}
\documentclass[runningheads]{llncs}

\usepackage[final,year=2026,ID=8705]{eccv}
\usepackage{eccvabbrv}
\usepackage{graphicx}
\usepackage{booktabs}
\usepackage{amsmath}
\usepackage{amssymb}
\usepackage{algorithm}
\usepackage{algorithmic}
\usepackage{wrapfig}
\restylefloat{algorithm}
\usepackage[accsupp]{axessibility}
\usepackage{xcolor}
\usepackage{hyperref}
\usepackage{orcidlink}
\usepackage{multirow}
\usepackage{xspace}

\begin{document}

\crefname{figure}{Fig.}{Figs.}
\Crefname{figure}{Figure}{Figures}
\crefname{equation}{Eq.}{Eqs.}
\Crefname{equation}{Equation}{Equations}
\crefname{section}{Sec.}{Secs.}
\Crefname{section}{Section}{Sections}
\crefname{algorithm}{Alg.}{Algs.}
\Crefname{algorithm}{Algorithm}{Algorithms}
\crefname{table}{Tab.}{Tabs.}
\Crefname{table}{Table}{Tables}
\crefname{appendix}{App.}{Apps.}
\Crefname{appendix}{Appendix}{Appendices}

\newcommand{\OM}{\textsc{GCA}\xspace}

\spnewtheorem{assumption}{Assumption}{\bfseries}{\itshape}

\title{Learning Implicit Constitutive Laws for Dynamic 3D Gaussian Splatting from Monocular Videos}

\titlerunning{Constitutive Alignment for Dynamic 3DGS}

\author{Xiaoyang Liu\inst{1}\orcidlink{0009-0005-5395-5710} \and
Kai Han\inst{1}\thanks{Corresponding author.}\orcidlink{0000-0002-7995-9999}}

\authorrunning{Xiaoyang Liu, Kai Han}

\institute{Visual AI Lab, The University of Hong Kong, Hong Kong \\
\email{xiaoyangliu@connect.hku.hk, kaihanx@hku.hk}}

\renewcommand{\thefootnote}{\fnsymbol{footnote}}
\setcounter{footnote}{1}
\maketitle
\renewcommand{\thefootnote}{\arabic{footnote}}
\setcounter{footnote}{0}

\begin{abstract}
We present \textsc{GCA}~(\textbf{G}aussian \textbf{C}onstitutive
\textbf{A}lignment), a framework for learning implicit constitutive
laws from monocular dynamic video of deformable objects represented by 3D Gaussians.
Given a static multi-view scan for geometric initialization, our method learns intrinsic physical dynamics solely from a single fixed-viewpoint video of the moving object.
Existing implicit methods often suffer from local minima under
noisy supervision and lack physical interpretability, while explicit
approaches rely on predefined constitutive equations, limiting
generalizability and becoming unstable in monocular settings.
To address these challenges, our
framework unifies LoRA-based adaptation with two key alignment modules. First, we propose
\textit{Rank-based Depth-Geometric Anchors} (RDGA) to establish robust
geometric constraints from monocular dynamic observations via
scale-invariant rank-based depth alignment, reducing the reliance on unreliable pixel-level color supervision. Second, a
\textit{Constitutive Prior Regularizer} (CPR) integrates classical
constitutive models as soft differentiable priors, regularizing the
optimization while preserving the flexibility of implicit
modeling---even when the actual material is absent from the
hypotheses. Extensive experiments on synthetic, real-to-sim, and real-world
datasets demonstrate that \OM outperforms existing methods, achieving
48\% lower Chamfer Distance than the strongest baseline on synthetic
benchmarks while remaining robust under monocular supervision.

\end{abstract}

\begin{figure*}[t]
  \centering
  \includegraphics[width=\textwidth]{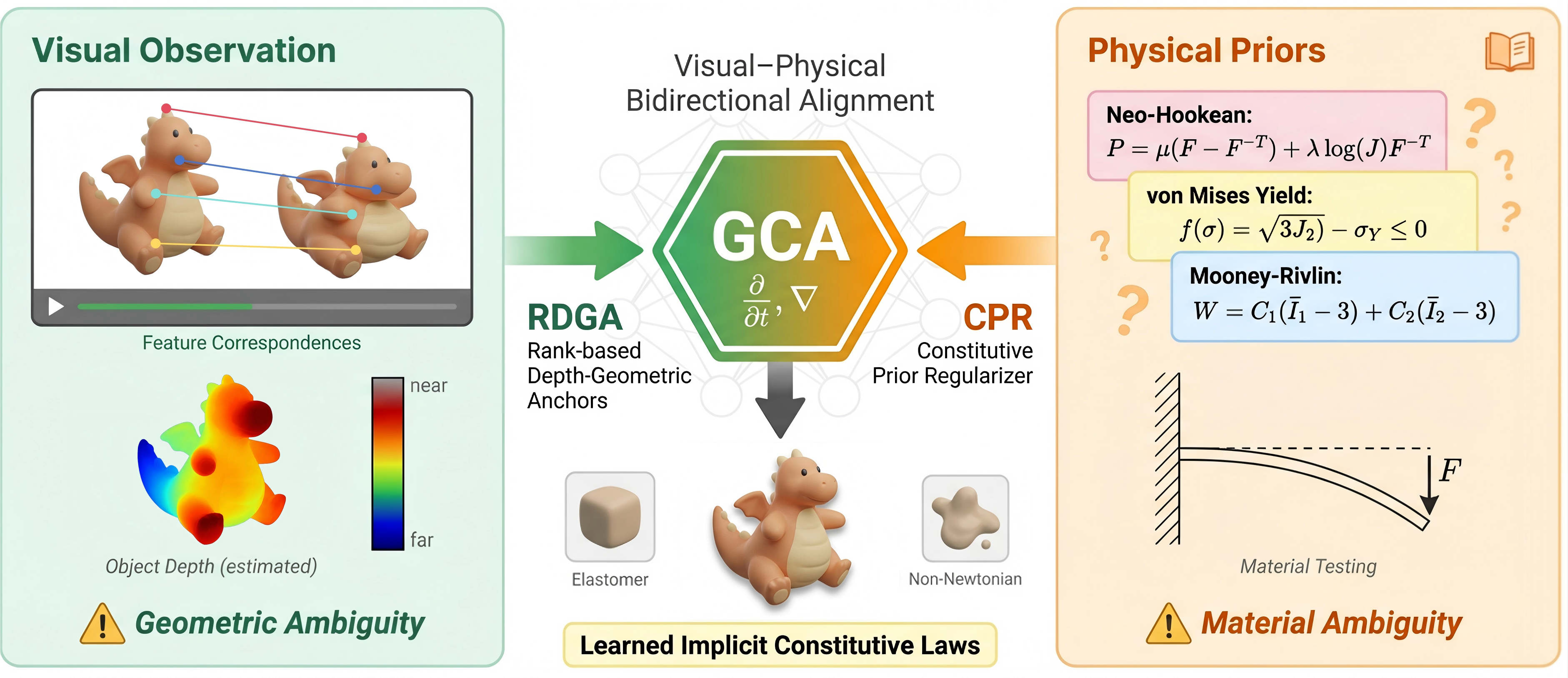}
  \caption{\textbf{The core idea of GCA.} Learning implicit constitutive laws from dynamic video faces two key challenges: \emph{geometric ambiguity} from sparse visual observations (left), and \emph{material ambiguity} from unknown constitutive models and parameters (right). GCA bridges both sides through visual--physical bidirectional alignment: Rank-based Depth-Geometric Anchors (RDGA) establish robust geometric constraints via scale-invariant depth consensus, while a Constitutive Prior Regularizer (CPR) softly guides optimization using classical constitutive hypotheses. Together, they enable reliable learning of implicit constitutive laws (bottom) that generalize across diverse material types.}
  \label{fig:teaser}
\end{figure*}

\section{Introduction}
\label{sec:intro}

Understanding the intrinsic dynamics of objects is crucial for spatial
intelligence, enabling accurate digital modeling, interaction, and
manipulation that follows physical
laws~\cite{yin2021modeling,digital1,robo1,robo2}. While humans
effortlessly infer basic physical properties from videos (\eg, bouncing
balls or viscous fluid flow), extracting precise physical models from
visual signals remains an open challenge.

Prior works have employed models to understand intrinsic dynamics
from visual
observations~\cite{chen2022virtual,jiang2024vr,huang2020learning,
huang2024dreamphysics,liu2024physgen}. A common approach combines
differentiable physics
simulators~\cite{dubied2022sim,xue2023jax} with differentiable
renderers~\cite{mildenhall2021nerf,kerbl20233d} for optimization.
Regarding constitutive law modeling, existing approaches fall into two
paradigms with complementary limitations:
\textit{Explicit modeling}~\cite{huang2024dreamphysics,liu2024physics3d,
li2023pac,zhang2024physdreamer} builds upon classical continuum
mechanics with predefined constitutive
models~\cite{drucker1952soil,von} (\eg, hyperelastic
models~\cite{stomakhin2012energetically}) and explicit parameters such
as Young's modulus and Poisson's ratio. While enabling physical
interpretability, effectiveness critically depends on correct model
specification: (1)~manual model selection and fine parameter tuning are
required for different materials~\cite{li2023pac}; (2)~complex
real-world materials with deeply coupled properties remain
intractable~\cite{liu2024physics3d}; and critically, (3)~these methods
perform poorly under monocular supervision, as observed in our experiments (\cref{tab:explicit_comparison}).
\textit{Implicit parameterization} models constitutive relations
through neural
networks~\cite{nclaw,li2022plasticitynet}. NeuMA~\cite{neuma}
introduces the first method to align implicit constitutive models with
visual observations. However, implicit models easily converge to
suboptimal solutions when handling noisy and sparse
supervision~\cite{wang2023softzoo}---a problem that becomes severe in
monocular settings.

These conflicting trade-offs lead to our core question: \textit{How to
reliably learn intrinsic dynamics from monocular video while preserving
the generalization advantages of implicit modeling?}
Following prior work in
dynamic 3D Gaussian splatting~\cite{springgaus,neuma}, we assume access to a
static multi-view orbital scan for initial 3D geometry reconstruction.
The focus of this work is to learn intrinsic constitutive laws from a single fixed-viewpoint video of the object undergoing dynamic deformation. This monocular dynamic setting introduces severe geometric
ambiguities (single viewpoint) and material ambiguities
(underconstrained physical parameters) that make the inverse problem
significantly ill-posed.

To address this challenge, we propose \OM, a framework that achieves
visual-physical bidirectional alignment for learning implicit
constitutive laws from monocular dynamic video. Our method unifies LoRA-based adaptation with two coordinated
alignment modules:
    (i)~Rank-based Depth-Geometric Anchors (\textbf{RDGA}) extract
    robust geometric constraints from sparse observations using a
    scale-invariant rank-based depth alignment mechanism. Unlike
    pixel-level color supervision, which suffers from domain shifts in
    monocular video, RDGA is designed to mitigate the scale-shift
    ambiguity of monocular depth estimators.
    (ii)~Constitutive Prior Regularizer (\textbf{CPR}) treats
    classical constitutive models as \emph{soft regularization priors}
    rather than hard constraints, resolving material ambiguity without
    sacrificing generalization, and remains effective even when the
    true material is absent from the hypotheses.

Extensive experiments validate that \OM achieves state-of-the-art
performance: 48\% lower Chamfer Distance than NeuMA on synthetic
benchmarks, robust performance on a challenging real-to-sim dataset,
and superior visual quality in real-world monocular experiments. We
also show that naively adapting explicit methods to monocular settings
leads to optimization divergence, highlighting the importance of
robust alignment.

\section{Related work}
\label{sec:related}

\noindent\textbf{Physics-grounded dynamic 3D generation.}
NeRF-based models~\cite{park2021nerfies,fang2022fast,lpd,pienerf} are
constrained by predefined material assumptions, while recent
Gaussian-based methods~\cite{kerbl20233d,splashing,physmotion} show
substantial progress; SpringGaus~\cite{springgaus} reconstructs elastic
dynamics but still relies on an explicit spring-mass model.
Diffusion-guided approaches~\cite{zhang2024physdreamer,liu2024physics3d,
huang2024dreamphysics,omniphysgs} inherit imprecise physics
priors~\cite{croitoru2023diffusion,dreamfusion,generative} and
typically assume rigid~\cite{liu2024physgen} or
elastic~\cite{springgaus} bodies. NeuMA~\cite{neuma} first optimizes
neural constitutive laws from observational images without predefined
laws, but single-modality visual optimization suffers from local minima
under sparse monocular supervision; naively augmenting explicit methods
(\eg, PAC-NeRF~\cite{li2023pac}) with monocular depth diverges,
motivating our hybrid alignment.

\noindent\textbf{Material constitutive laws.}
Conventional approaches~\cite{arruda1993three,von,chhabra2023bubbles,
gic,liu2024physics3d} enforce explicit laws via predefined nonlinear
polynomial bases (\eg, elastic~\cite{fung1967elasticity}/plastic~%
\cite{drucker1952soil}/fluid~\cite{chhabra2023bubbles} models) and
tune parameters such as Young's modulus or Poisson's ratio, requiring
manual specification~\cite{survey0}. Implicit neural
modeling~\cite{raissi2019pinn,pinns,deeponet} bypasses this:
NCLaw~\cite{nclaw} pioneers hybrid NN--PDE training but requires
particle-level annotations; NeuMA~\cite{neuma} avoids particle ground
truth via LoRA-based~\cite{hu2022lora} alignment with differentiable
rendering, yet pure visual supervision lacks physical
interpretability~\cite{aira2024motioncraft} and sparse observations
introduce optimization ambiguity. Our framework softly regularizes
implicit optimization with physical priors, unifying the stability of
explicit laws with the generalization of implicit ones under sparse
supervision.

\section{Method}

\subsection{Problem statement}
Given static 3D Gaussians~\cite{kerbl20233d} of an object
$\mathcal{G}(i) = \{\mathbf{p}(i), \alpha(i), \mathbf{A}(i),
\mathbf{c}(i)\}$, where $\mathbf{p}(i), \alpha(i), \mathbf{A}(i),
\mathbf{c}(i)$ are the center, opacity, covariance matrix, and
spherical harmonic coefficients of each Gaussian primitive, and its
corresponding monocular dynamic video $\{I_t\}_{t=1}^T$, we aim to
learn implicit constitutive laws through a dynamical system
$\mathcal{M}_\theta$ governed by elastoplastic dynamics~\cite{fung1977first}:
\begin{equation}
\label{equation:elasto}
    \rho_0 \ddot{\phi} = \nabla \cdot \mathbf{P} + \rho_0 \mathbf{b},
    \quad
    \mathbf{P} = \mathcal{E}(\mathbf{F}_e),\quad
    \mathbf{F}_e = \nabla\phi,
\end{equation}
where $\mathbf{P}$ is the first Piola--Kirchhoff stress tensor,
$\rho_0$ is the object density, and $\mathbf{b}$ is the body force.
$\phi$ denotes the deformation map, $\ddot{\phi}$ is its acceleration,
and $\mathcal{E}$ is the elastic constitutive law.

We discretize~\cref{equation:elasto} and obtain the dynamical system
$\mathcal{M}_\theta$:
\begin{equation}
\label{equation:transition}
    \mathbf{s}_{t+1} = \mathcal{M}_\theta(\mathbf{s}_t),\quad
    \forall\, t= 0,1,\dots,T-1,
\end{equation}
where $\mathbf{s}_t = \{\mathbf{x}_t, \mathbf{v}_t, \mathbf{F}_e^t\}$
denotes the state at time step $t$, consisting of particle positions,
velocities, and elastic deformation gradients. $\theta$ denotes the
neural parameters in $\mathcal{M}$. Additional details on preprocessing Gaussians to particles are provided in the supplementary material.

\begin{figure}[t]
    \centering
    \includegraphics[width=\linewidth]{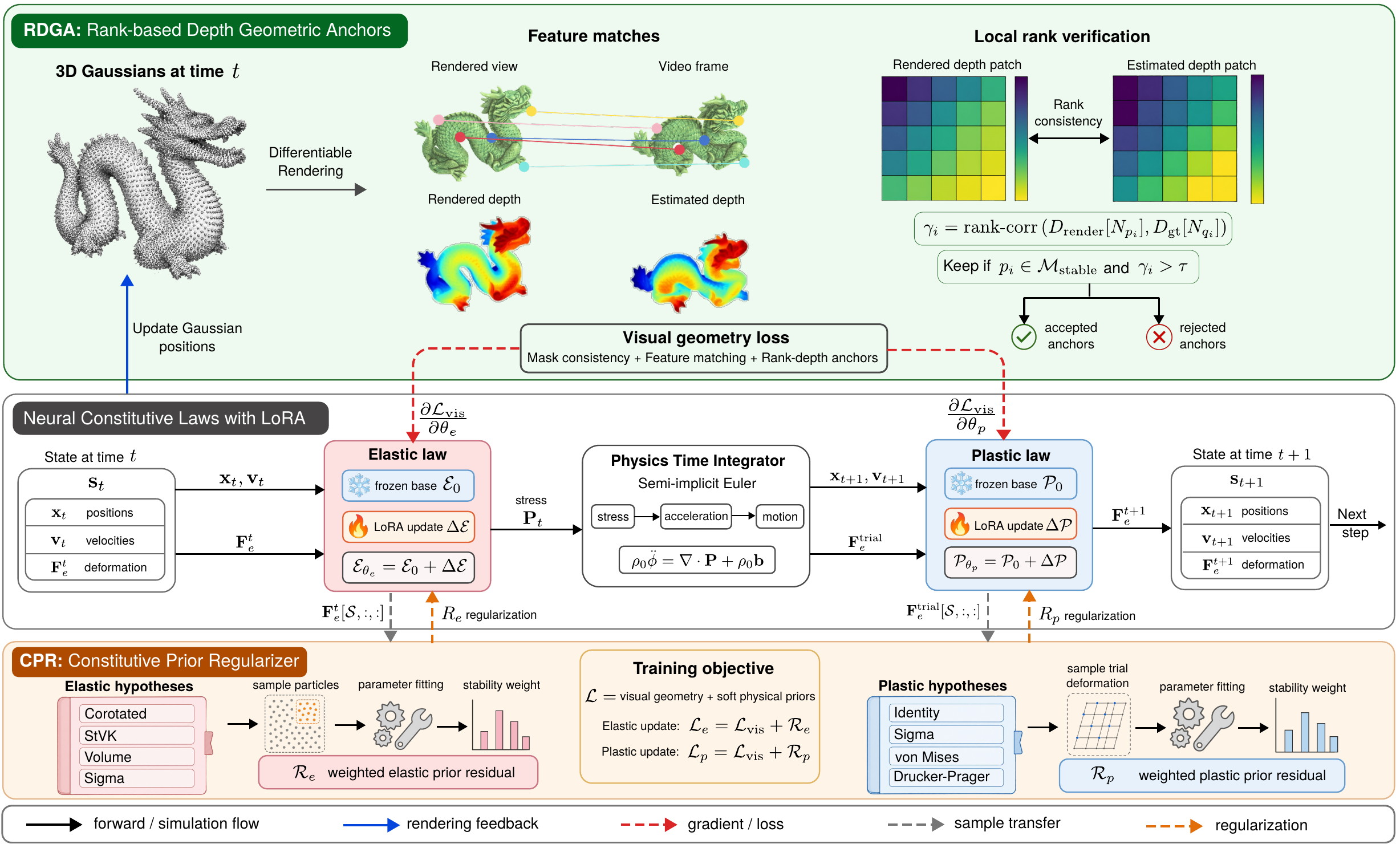}
    \caption{\textbf{Overview of \OM.} Our framework builds on
    Low-Rank Adaptation (LoRA) to fine-tune neural material laws while
    maintaining integration with PDE-based simulation, and introduces
    two coordinated alignment modules: (i)~Rank-based Depth-Geometric
    Anchors (RDGA) that resolve geometric ambiguities via
    scale-invariant rank-based depth alignment, and (ii)~a
    Constitutive Prior Regularizer (CPR) that provides soft physical
    priors while preserving generalization.}
    \label{fig:method}
\end{figure}

To align $\mathcal{M}_\theta$ with observation $I_t$, we use a
differentiable 3DGS renderer $\mathcal{R}$ producing
$\hat{I}_t = \mathcal{R}(\mathbf{s}_t; \mathbf{K}, \mathbf{Q})$,
where $\mathbf{K}, \mathbf{Q}$ denote camera intrinsic and extrinsic
matrices. Relying solely on this rendering-based supervision, however,
is insufficient to overcome the challenges posed by sparse and noisy
monocular video. The inherent \textit{geometric ambiguities} from the
single viewpoint and \textit{material ambiguities} in the dynamics make
the optimization landscape intractable. To establish a robust learning
pipeline, we propose \OM (see~\cref{fig:method}), which combines
LoRA-based adaptation with two coordinated alignment modules---%
Rank-based Depth-Geometric Anchors (\textbf{RDGA}) and a Constitutive
Prior Regularizer (\textbf{CPR})---detailed below.

\subsection{Neural material constitutive laws}
\label{sec:nclaw}
Our work adopts the same dynamical system $\mathcal{M}_\theta$ as
NCLaw~\cite{nclaw} for state transitions. $\mathcal{M}_\theta$ is
composed of the neural elasticity law $\mathcal{E}_{\theta_e}$,
semi-implicit Euler integration~\cite{mpm3,mpm2}, and neural plasticity law
$\mathcal{P}_{\theta_p}$. We use the basic physical prior model
$\mathcal{M}_0 = \{\mathcal{E}_0, \mathcal{P}_0\}$ provided by NCLaw.
To align the model with observations without compromising its
fundamental capabilities, we use LoRA~\cite{hu2022lora} for
fine-tuning instead of training all parameters. Specifically,
$\mathcal{M}_\theta = \{\mathcal{E}_{\theta_e},
\mathcal{P}_{\theta_p}\}$, where
$\mathcal{E}_{\theta_e} = \mathcal{E}_0 + \Delta
\mathcal{E}_{\theta_e}$ and
$\mathcal{P}_{\theta_p} = \mathcal{P}_0 + \Delta
\mathcal{P}_{\theta_p}$.

\noindent\textbf{Remark.} While our dynamical system builds upon the
NCLaw architecture~\cite{nclaw}, NCLaw itself requires
\emph{dense particle-level ground-truth supervision}, which is
unavailable in visual observation settings. The core challenge
addressed by \OM is not the design of neural constitutive
architectures, but the \emph{visual-physical alignment} problem: how
to provide reliable supervisory signals to train such architectures
from noisy, sparse, single-viewpoint video. As shown
in~\cref{tab:explicit_comparison}, naively adapting existing methods
to monocular settings leads to optimization instability, highlighting the
importance of robust alignment. Moreover, as
demonstrated in~\cref{tab:efficiency}, LoRA not only improves parameter
efficiency but also acts as a regularizer---full fine-tuning tends to
overfit visual noise, yielding worse physical accuracy.

\begin{figure}[t]
\centering
\begin{minipage}{0.7\linewidth}
\begin{algorithm}[H]
\caption{Time stepping}
\label{alg:simulation}
\small
\centering
\begin{minipage}{\dimexpr\linewidth-1.5em\relax}
\begin{algorithmic}
\REQUIRE State $\mathbf{s}_t = \{\mathbf{x}_t, \mathbf{v}_t,
\mathbf{F}_e^t\}$
\ENSURE Next state $\mathbf{s}_{t+1}$
\STATE \textbf{Stress Evaluation:}
\FOR{each material point $i=1$ to $N$}
\STATE $\mathbf{P}_t^{(i)} \gets
\mathcal{E}_{\theta_e}(\mathbf{F}_e^{t,(i)}; \theta_e)$
\ENDFOR
\STATE \textbf{Euler Integration:}
\STATE $\mathbf{x}_{t+1}, \mathbf{v}_{t+1},
\mathbf{F}_e^{\text{trial}} \gets
\mathcal{I}(\mathbf{x}_t, \mathbf{v}_t, \mathbf{P}_t)$
\STATE \textbf{Plasticity Update:}
\FOR{each material point $i=1$ to $N$}
\STATE $\mathbf{F}_e^{t+1,(i)} \gets
\mathcal{P}_{\theta_p}(\mathbf{F}_e^{\text{trial},(i)}; \theta_p)$
\ENDFOR
\end{algorithmic}
\end{minipage}
\end{algorithm}
\end{minipage}
\end{figure}
The dynamical system $\mathcal{M}_\theta$ advances physical states
through three stages as shown in~\cref{alg:simulation}:
(1)~\textit{Stress evaluation} via neural constitutive law
$\mathcal{E}_{\theta_e}$ that computes first Piola--Kirchhoff stress
$\mathbf{P}_t$ from elastic deformation gradient $\mathbf{F}_e^t$;
(2)~\textit{Dynamics integration} where operator $\mathcal{I}$
implements semi-implicit Euler integration to update positions
$\mathbf{x}_{t+1}$ and velocities $\mathbf{v}_{t+1}$;
(3)~\textit{Plasticity update} through network
$\mathcal{P}_{\theta_p}$ that modifies $\mathbf{F}_e^{\text{trial}}$
to account for plastic deformation.

\subsection{Rank-based depth-geometric anchors}
\label{sec:rdga}

Pixel-level color matching, adopted by methods like NeuMA~\cite{neuma}, struggles under monocular video due to lighting
variations, self-occlusions during rapid rotation, and geometric ambiguities. To overcome this, we propose Rank-based
Depth-Geometric Anchors (RDGA), which establish robust geometric constraints by enforcing rank-based consensus between rendered depth
and estimated depth, focusing on stable interior regions.

\noindent\textbf{Design rationale.} Monocular depth estimators suffer
from inherent scale and shift ambiguity---absolute depth values are
unreliable across frames. Standard metric losses (L1/L2) amplify this
noise, as validated in~\cref{tab:detailed_ablation} where replacing
our rank-based formulation with L1 loss degrades performance below the
baseline. Our Spearman rank correlation is scale-invariant: it only
requires that relative depth orderings are preserved, which modern
estimators achieve reliably even under large deformations.

Given a rendered image $I_{\text{render}} \in \mathbb{R}^{H \times W
\times 3}$, a rendered depth map
$D_{\text{render}} \in \mathbb{R}^{H \times W}$, and a ground-truth
(GT) image $I_{\text{gt}} \in \mathbb{R}^{H \times W \times 3}$, we first compute a silhouette consistency loss:
\begin{equation}
\mathcal{L}_{\text{mask}} = \|M_{\text{render}} - M_{\text{gt}}\|_2^2,
\end{equation}
where $M_{\text{render}}$ and $M_{\text{gt}}$ are object region masks
on rendered and GT images.

We utilize a pre-trained depth estimation network $\mathcal{D}$
(\eg, DAV~\cite{dav}) to generate relative depth maps
$D_{\text{gt}}$. A feature matching network $\mathcal{F}$
(\eg, SuperGlue~\cite{superglue,superpoint}) extracts $N$ 2D
correspondence pairs $\{(p_i, q_i)\}_{i=1}^N$, where
$p_i = (x_i, y_i)$ and $q_i = (x'_i, y'_i)$ denote matched
coordinates in $I_{\text{render}}$ and $I_{\text{gt}}$. To mitigate
depth estimation errors near object boundaries, we define the edge
region $\mathcal{M}_{\text{edge}}$ through morphological opening:
\begin{equation}
\mathcal{M}_{\text{edge}} = \mathcal{M} \circ K_{r\times r},
\end{equation}
where $\circ$ denotes morphological opening with an $r \times r$
rectangular kernel $K$. The stable interior region is
$\mathcal{M}_{\text{stable}} = \mathcal{M} \setminus
\mathcal{M}_{\text{edge}}$.

For each correspondence pair $(p_i, q_i)$, we validate depth consensus
in local neighborhoods. Let $N_{p_i}$ and $N_{q_i}$ represent
$k \times k$ regions centered at $p_i$ in $D_{\text{render}}$ and
$q_i$ in $D_{\text{gt}}$. We compute the Spearman rank correlation
coefficient $\gamma_i$~\cite{spearman}:
\begin{equation}
\gamma_i = 1 - \frac{6 \sum_{j=1}^{k^2} (r_j - s_j)^2}
{k^2(k^4 - 1)},
\end{equation}
where $r_j$ and $s_j$ are the ranks of the $j$-th depth value in
$N_{p_i}$ and $N_{q_i}$. A consensus indicator function filters pairs:
\begin{equation}
\phi(p_i, q_i) = \mathbb{I}(\gamma_i > \tau).
\end{equation}
Only pairs in $\mathcal{M}_{\text{stable}}$ satisfying
$\phi(p_i, q_i) = 1$ are retained:
$\mathcal{A} = \{(p_i, q_i) \mid p_i \in \mathcal{M}_{\text{stable}}
\land \phi(p_i, q_i) = 1\}$.

The geometric alignment objective combines global and anchor-level
terms:
\begin{equation}
\mathcal{L}_{\text{geo}} = \underbrace{\lambda_1 \left\| P_{\text{render}} -
P_{\text{gt}} \right\|_2}_{\text{global alignment}} +
\underbrace{\lambda_2 \sum_{(p_i, q_i) \in \mathcal{A}}
(-\gamma_i)}_{\text{anchor-level supervision}},
\end{equation}
where $P_{\text{render}}$ and $P_{\text{gt}}$ represent matched point
sets. The first term maintains global geometric consistency while the
second focuses on precise local relationships through verified anchor
pairs.

\subsection{Constitutive prior regularizer}
\label{sec:cpr}

To resolve material ambiguity in sparse-view settings, we design a
constitutive prior regularization process that evaluates candidate
constitutive laws through parameter stability analysis.

\noindent\textbf{Constitutive hypotheses.} Following NCLaw~\cite{nclaw}, our elastic hypothesis set
$\mathcal{H}_e$ includes four models: (1)~corotated elasticity,
(2)~St.~Venant--Kirchhoff (StVK) elasticity, (3)~volume elasticity,
and (4)~sigma elasticity. The plastic hypothesis set $\mathcal{H}_p$
also includes four models: (1)~identity plasticity,
(2)~sigma plasticity, (3)~von Mises plasticity~\cite{von},
and (4)~Drucker--Prager plasticity~\cite{drucker1952soil}.
These candidates cover common elastic, volumetric, and plastic
response families. Full formulations are provided in the supplementary material. Importantly, CPR acts as a \emph{soft regularization prior}---it
remains effective even when the true material behavior does not match
any of the hypotheses, as validated in~\cref{tab:ood}.

The elastic deformation gradient
$\mathbf{F}_e^t \in \mathbb{R}^{N\times3\times3}$ serves as input,
with physics residuals $\mathcal{R}_e, \mathcal{R}_p$ quantifying
deviation from plausible laws. For each candidate law, we use
$\mathbf{F}_e^t[\mathcal{S},:,:]$ to evaluate alignment, where
$\mathcal{S} \subset \{1,\dots,N\}$ is a fixed subset of indices
($|\mathcal{S}|=256$ by default).

The Constitutive Prior Regularizer
(\cref{alg:phys_sim}) operates through three phases. First, it
performs standard elastic-plastic simulation: elastic stress prediction
via $\mathcal{E}_{\theta_e}$, Euler integration through $\mathcal{I}$,
and plasticity correction using $\mathcal{P}_{\theta_p}$. Next, it
solves inverse problems to estimate explicit parameters $\Theta$ for
each candidate law. The credibility weights $\omega$ are computed as
inverse variance measures (with smoothing factor $\epsilon$), assigning
higher confidence to laws with stable parameter estimates. Finally,
physical residuals $\mathcal{R}_e$ and $\mathcal{R}_p$ penalize
deviations using weighted combinations.

\begin{figure}[t]
\centering
\begin{minipage}{0.85\linewidth}
\begin{algorithm}[H]
\caption{Constitutive prior regularizer}
\label{alg:phys_sim}
\small
\centering
\begin{minipage}{\dimexpr\linewidth-1.5em\relax}
\begin{algorithmic}
\REQUIRE
\STATE $\mathbf{F}_e^t \in \mathbb{R}^{N\times3\times3}$,
$\mathcal{H}_e = \{\mathcal{H}_e^k\}_{k=1}^K$,
$\mathcal{H}_p = \{\mathcal{H}_p^m\}_{m=1}^M$,
$\mathcal{E}_{\theta_e}$, $\mathcal{P}_{\theta_p}$, $\epsilon$,
$\mathcal{S} \subset \{1,\dots,N\}$
\ENSURE
\STATE $\mathbf{F}_e^{t+1}$, $\mathcal{R}_e^t$, $\mathcal{R}_p^t$
\STATE \textbf{Elastic Stress Prediction:}
$\mathbf{P}^t \gets \mathcal{E}_{\theta_e}(\mathbf{F}_e^t)$
\STATE \textbf{Euler Integration:}
$\mathbf{F}_e^{\text{trial}} \gets \mathcal{I}(\mathbf{P}^t)$
\STATE \textbf{Plasticity Correction:}
$\mathbf{F}_e^{t+1} \gets
\mathcal{P}_{\theta_p}(\mathbf{F}_e^{\text{trial}})$
\STATE \textbf{Parameter Solving:}
\STATE $\mathbf{F}_e^{t,\mathcal{S}} \gets
\mathbf{F}_e^t[\mathcal{S},:,:]$; \quad
$\mathbf{F}_e^{\text{trial},\mathcal{S}} \gets
\mathbf{F}_e^{\text{trial}}[\mathcal{S},:,:]$
\FOR{$k=1$ \TO $K$}
\STATE $\hat{\Theta}_e^{k} \gets
\mathop{\arg\min}\limits_{\Theta_e^k}
\|\mathcal{E}_{\theta_e}(\mathbf{F}_e^{t,\mathcal{S}}) -
\mathcal{H}^k_e(\mathbf{F}_e^{t,\mathcal{S}}; \Theta_e^k)\|_F^2$
\STATE $\omega_e^k \gets
\frac{1}{\text{Var}\{\hat{\Theta}_e^k\} + \epsilon}$
\ENDFOR
\FOR{$m=1$ \TO $M$}
\STATE $\hat{\Theta}_p^{m} \gets
\mathop{\arg\min}\limits_{\Theta_p^m}
\|\mathcal{P}_{\theta_p}(\mathbf{F}_e^{\text{trial},\mathcal{S}}) -
\mathcal{H}^m_p(\mathbf{F}_e^{\text{trial},\mathcal{S}};
\Theta_p^m)\|_F^2$
\STATE $\omega_p^m \gets
\frac{1}{\text{Var}\{\hat{\Theta}_p^m\} + \epsilon}$
\ENDFOR
\STATE $\mathcal{R}_e^t \gets \sum_{k=1}^K \omega_e^k
\|\mathcal{E}_{\theta_e}(\mathbf{F}_e^t) -
\mathcal{H}_e^k(\mathbf{F}_e^t;
\mathbb{E}[\hat{\Theta}_e^{k}])\|_F^2$
\STATE $\mathcal{R}_p^t \gets \sum_{m=1}^M \omega_p^m
\|\mathcal{P}_{\theta_p}(\mathbf{F}_e^{\text{trial}}) -
\mathcal{H}_p^m(\mathbf{F}_e^{\text{trial}};
\mathbb{E}[\hat{\Theta}_p^{m}])\|_F^2$
\end{algorithmic}
\end{minipage}
\end{algorithm}
\end{minipage}
\end{figure}

\noindent\textbf{Overall optimization objective.} The total loss for
optimizing the neural elastic model $\mathcal{E}_{\theta_e}$ and neural
plasticity model $\mathcal{P}_{\theta_p}$ is
\begin{equation}
\mathcal{L}=\lambda_m\mathcal{L}_{\text{mask}}+
\lambda_g\mathcal{L}_{\text{geo}}+\mathcal{R}_e+\mathcal{R}_p,
\end{equation}
where $\lambda_m$ and $\lambda_g$ are balance factors. A theoretical
analysis of the convergence properties is provided in the supplementary material.

\section{Experiments}

\subsection{Experimental setup}
\label{sec:exp_setup}

\noindent\textbf{Datasets.} We conduct systematic validation across
three dimensions. For \textit{synthetic} experiments, we construct a
challenging benchmark of six material types (elastomers, gels, rubber,
plasticine, granular materials, and non-Newtonian fluids) across
diverse object geometries; the benchmark introduces (1)~randomized
lighting interference, (2)~reduced frame rates matching real-world
constraints, and (3)~compound material modeling, with additional details
provided in the supplementary material. For the \textit{real-to-sim} bridge, we
capture high-quality 3D Gaussian splatting models of real objects
(\textit{dragon}, \textit{wolf}, \textit{pudding}) and simulate dynamic
sequences with complex material properties, providing ground-truth
physics while retaining real-world geometric complexity. For
\textit{real-world} validation, we adopt the
SpringGaus~\cite{springgaus} dataset; while its original setup employs
tri-view supervision, we strictly constrain our approach to single-view
video, consistent with our problem setting. All experiments run on a
single NVIDIA A800 80GB GPU.

\noindent\textbf{Baseline methods.} We evaluate against:
(1)~NCLaw~\cite{nclaw} (implicit, requires particle GT);
(2)~NeuMA~\cite{neuma} (implicit, visual supervision);
(3)~SpringGaus~\cite{springgaus} (explicit, real-world); and
(4)~PAC-NeRF~\cite{li2023pac} (explicit, multi-view), which
we additionally adapt to monocular settings by augmenting with
monocular depth supervision. We
exclude PhysDreamer~\cite{zhang2024physdreamer} and
Physics3D~\cite{liu2024physics3d} as their reliance on diffusion
guidance and predefined explicit models is orthogonal to our implicit
learning objective.

\noindent\textbf{Evaluation metrics.}
We use: (1)~Chamfer Distance (CD)~\cite{cd,cd2} for geometric
consistency, (2)~SSIM~\cite{ssim} for structural similarity,
(3)~PSNR~\cite{psnr} for pixel-level accuracy, and
(4)~LPIPS~\cite{lpips} for perceptual similarity.

\subsection{Evaluation on synthetic dataset}

We evaluate physical simulation accuracy by computing Chamfer Distance
between predicted and ground-truth particle positions.
\Cref{tab:chamfer_distance} shows that under our challenging benchmark
with color inconsistency and sparse supervision, our method achieves
48\% lower average CD than NeuMA.

\begin{table}[H]
    \centering
    \caption{\textbf{Quantitative comparison on synthetic dataset
    (Chamfer Distance $\downarrow$).} Our method achieves 48\% lower
    average CD than NeuMA across diverse materials.}
    \label{tab:chamfer_distance}
    \resizebox{\linewidth}{!}{%
    \begin{tabular}{lccccccc}
        \toprule
        Material & Elastomer & Gel & Rubber & Plasticine & Granular
        & Non-Newt. & \\
        Object & Ball & Duck & Pawn & Cat & Fish & Bottle & Average \\
        \midrule
        NCLaw~\cite{nclaw} & 4.085 & 2.934 & 2.031 & 1.909 & 0.536
        & 1.631 & 2.188 \\
        NeuMA~\cite{neuma} & 1.123 & 1.863 & 0.517 & 0.844 & 0.322
        & 1.056 & 0.954 \\
        \textbf{Ours} & \textbf{0.922} & \textbf{0.702} &
        \textbf{0.200} & \textbf{0.318} & \textbf{0.077} &
        \textbf{0.757} & \textbf{0.496} \\
        \bottomrule
    \end{tabular}}
\end{table}

\begin{figure}[t]
    \centering
    \includegraphics[width=\linewidth]{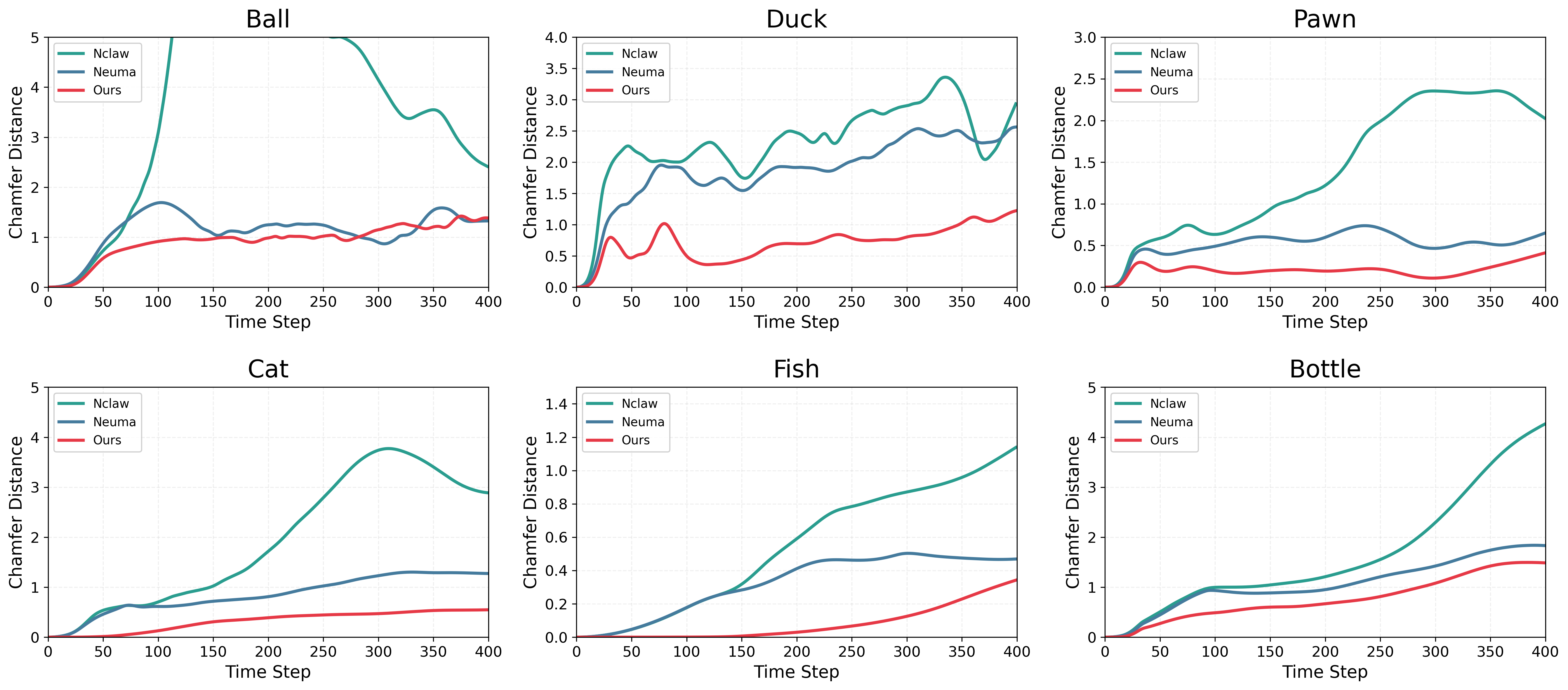}
    \caption{\textbf{Chamfer Distance during physical simulation.} Our
    method consistently maintains lower CD throughout the simulation.}
    \label{fig:cdnpy}
\end{figure}

\begin{figure}[H]
    \centering
    \includegraphics[width=\linewidth]{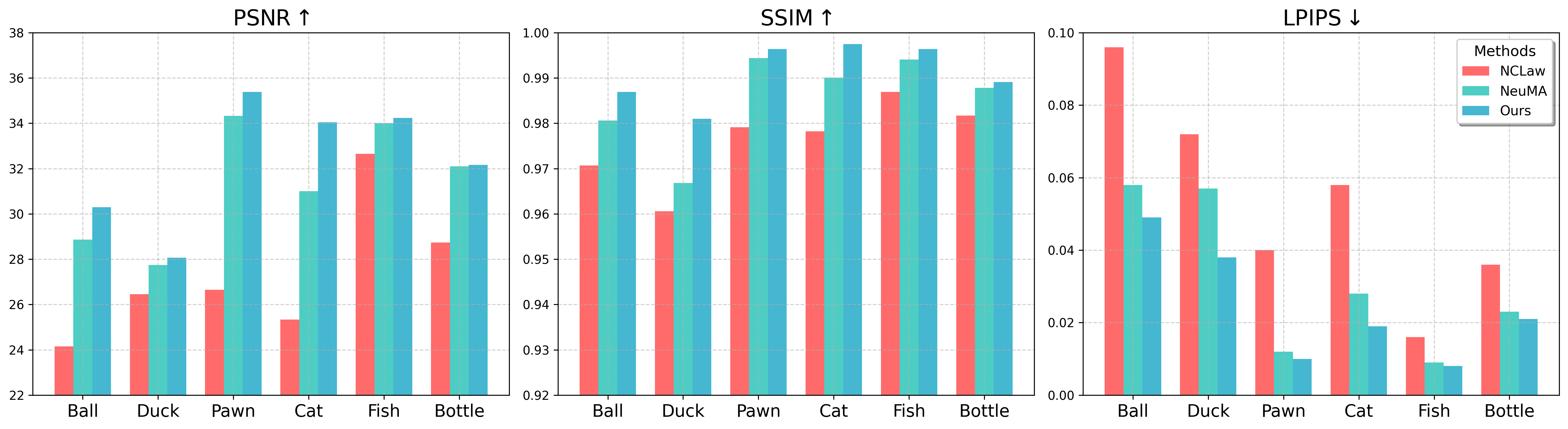}
    \caption{\textbf{Rendering metrics on synthetic dataset.} Our
    method achieves superior PSNR, SSIM, and LPIPS.}
    \label{fig:render_metrics}
\end{figure}

\Cref{fig:cdnpy} illustrates temporal CD variations during simulation.
Our method maintains alignment with ground truth throughout the
simulation, while baselines gradually deviate. Further rendering metrics
are provided in~\cref{fig:render_metrics}.

\subsection{Evaluation on real-to-sim dataset}

\begin{figure}[H]
    \centering
    \includegraphics[width=\linewidth]{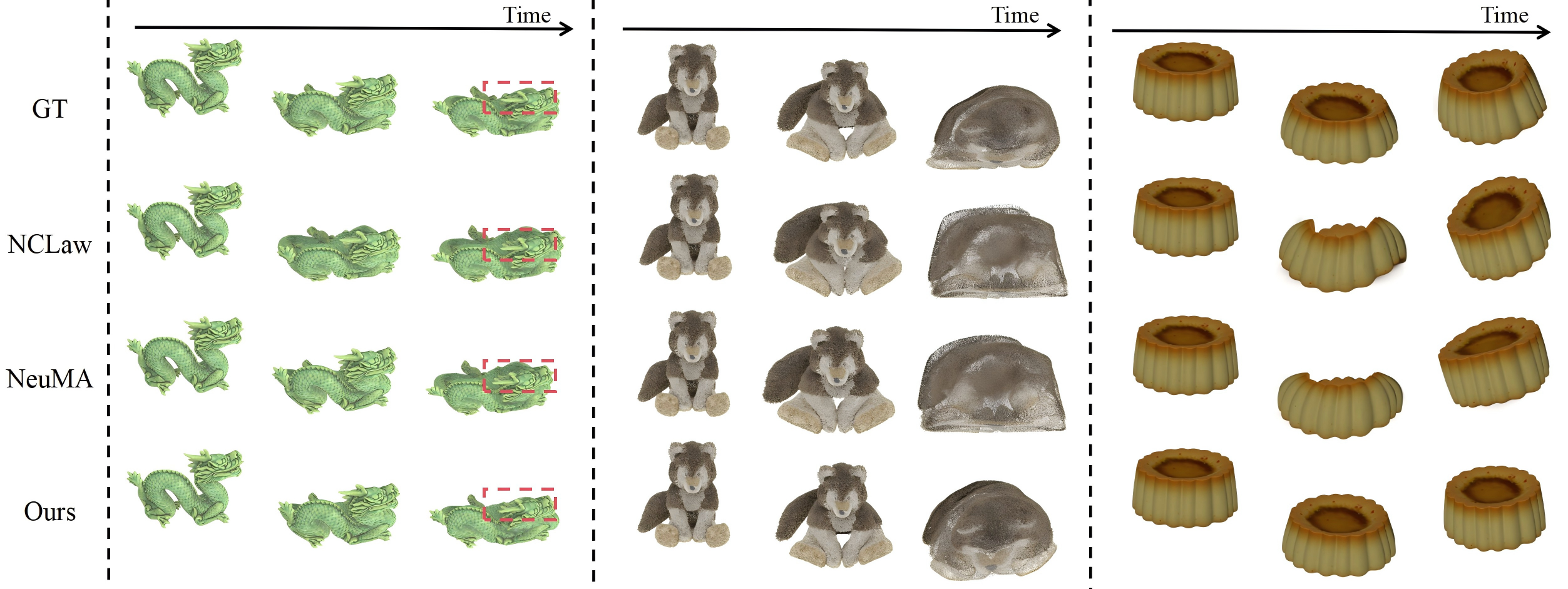}
    \caption{\textbf{Qualitative comparison on real-to-sim dataset.}
    Our method accurately captures complex dynamics for real-world
    geometries.}
    \label{fig:realtosim_quality}
\end{figure}

We assess generalization to complex geometries derived from real
objects using our real-to-sim dataset. \Cref{tab:realtosim_chamfer}
shows \OM consistently outperforms baselines.
\Cref{fig:realtosim_quality} provides qualitative comparison,
demonstrating superior ability to capture plausible dynamics for
complex objects.

\begin{table}[H]
    \centering
    \caption{\textbf{Chamfer Distance on real-to-sim dataset
    ($\downarrow$).} Our method achieves the lowest error across all
    objects.}
    \label{tab:realtosim_chamfer}
    \begin{tabular}{lccc}
        \toprule
        Object & Plast.\ Dragon & Sand Wolf & Gel Pudding \\
        \midrule
        NCLaw~\cite{nclaw} & 25.021 & 49.420 & 38.149 \\
        NeuMA~\cite{neuma} & 3.527 & 9.803 & 13.804 \\
        \textbf{Ours} & \textbf{2.081} & \textbf{5.842} &
        \textbf{0.906} \\
        \bottomrule
    \end{tabular}
\end{table}

\subsection{Evaluation on real-world dataset}

\begin{figure}[H]
    \centering
    \includegraphics[width=\linewidth]{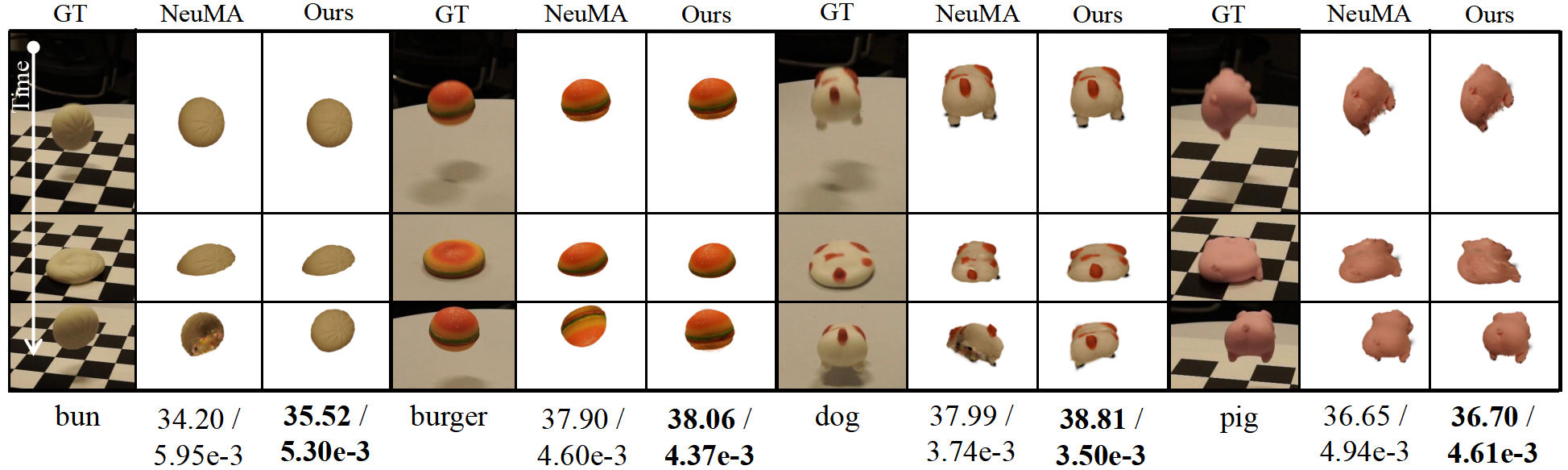}
    \caption{\textbf{Qualitative comparison on real-world dataset.} Our
    method achieves better visual consistency using only monocular supervision.
    The PSNR and LPIPS values are computed after filtering the background.}
    \label{fig:gaus}
\end{figure}

For real-world validation, we conduct monocular supervision experiments
on the SpringGaus dataset~\cite{springgaus}. While the original setup
uses tri-view supervision, we strictly use single-view video. As shown
in~\cref{fig:gaus}, \OM successfully disentangles implicit physical
properties and demonstrates strong generalization under monocular
constraints.

\subsection{Ablation and analysis}
\label{sec:ablation}

We conduct comprehensive ablation studies to validate each component
and design choice.

\noindent\textbf{Component-wise ablation.}
\Cref{tab:detailed_ablation} provides a granular breakdown of each
component's contribution.

\begin{table}[H]
    \centering
    \caption{\textbf{Detailed ablation study (Chamfer Distance
    $\downarrow$).} Each component contributes meaningfully. Replacing
    rank-based depth alignment with L1 loss degrades performance
    \emph{below the baseline}, validating our scale-invariant design.}
    \label{tab:detailed_ablation}
    \resizebox{\linewidth}{!}{%
    \begin{tabular}{lccccccl}
        \toprule
        Configuration & Elast. & Gel & Rubber & Plast. & Gran.
        & Non-Newt. & Note \\
        \midrule
        NeuMA (Baseline) & 1.123 & 1.863 & 0.517 & 0.844 & 0.322
        & 1.056 & Prior best \\
        \midrule
        w/o Global Align. & \multicolumn{6}{c}{$>100$ (Diverged)}
        & Divergence \\
        w/o Anchor Superv. & 1.024 & 0.963 & 0.637 & 0.510 & 0.164 & 1.166 & Loss of detail \\
        Replace Rank w/ L1 & 1.064 & 1.098 & 0.846 & 0.919 & 0.147 & 0.989 & Scale ambiguity \\
        w/o RDGA & 3.842 & 3.045 & 1.975 & 1.821 & 0.559 & 1.531
        & No geom.\ anchor \\
        w/o CPR & 1.024 & \textbf{0.694} & 0.243 & 0.510
        & 0.084 & 0.769 & No physics prior \\
        \midrule
        \textbf{Ours (Full)} & \textbf{0.922} & 0.702 &
        \textbf{0.200} & \textbf{0.318} & \textbf{0.077} &
        \textbf{0.757} & Best performance \\
        \bottomrule
    \end{tabular}}
\end{table}

\noindent(1)~\textit{Global alignment is foundational}: removing it
causes complete optimization divergence, as the model cannot capture
overall motion trends.
(2)~\textit{Rank-based $>$ Metric-based}: replacing our rank-based
loss with standard L1 depth loss yields CD of 0.844 (avg.),
significantly worse than our full model (0.496) and showing
\emph{degraded performance on 5 of 6 materials} compared to our
rank-based formulation. This is because monocular depth estimators
suffer from scale-shift ambiguity---forcing absolute scale match
introduces noise rather than valid supervision. Our rank-based
consensus mitigates this issue.
(3)~\textit{RDGA is critical}: without it, performance degrades
sharply across all materials.
(4)~\textit{CPR provides consistent benefit}: the regularizer improves
performance on 5 of 6 materials. The single exception (Gel: 0.694
vs.\ 0.702) is within noise margins and reflects that simple elastic
materials may not require additional physics guidance.
(5)~\textit{Color supervision is unreliable under monocular shading.}
In monocular videos of dynamic objects, self-occlusions and rapid
rotation cause severe lighting
changes and shading artifacts. Geometric consistency via depth provides
a more invariant signal for learning physical dynamics.

\noindent\textbf{Computational efficiency.}
\Cref{tab:efficiency} shows LoRA reduces GPU memory by 60\% and training time by 87\% versus full fine-tuning, while achieving superior CD (0.50 vs.\ 1.35). Full fine-tuning overfits visual noise.

\begin{table}[H]
    \centering
    \caption{\textbf{Computational efficiency analysis.} LoRA
    fine-tuning achieves superior physical accuracy with lower cost.}
    \label{tab:efficiency}
    \begin{tabular}{lcccc}
        \toprule
        Method & Mem.\ (GB) & Train (min) & Infer.\ (s) $\downarrow$ & CD $\downarrow$ \\
        \midrule
        NeuMA~\cite{neuma} & 28.9 & 73.3 & 32.4 & 1.31 \\
        Ours (Full F.T.) & 76.9 & 408.6 & 32.5 & 1.35 \\
        \textbf{Ours (LoRA)} & \textbf{30.5} & \textbf{51.2} & 31.5
        & \textbf{0.50} \\
        \bottomrule
    \end{tabular}
\end{table}

\noindent\textbf{Explicit methods are unstable under monocular supervision.}
A natural question is whether existing explicit methods can be adapted
to monocular settings by adding depth supervision.
\Cref{tab:explicit_comparison} provides empirical evidence: PAC-NeRF
augmented with monocular depth becomes unstable and diverges in our monocular setting (CD $>100$).
The rigid parameterization of explicit models (\eg, Young's modulus)
makes them overly sensitive to geometric noise inherent in monocular
depth estimates. These results suggest that directly adding monocular depth supervision
to existing explicit frameworks is insufficient---the specific design
of RDGA (rank-based, edge-aware) and CPR (soft regularization) is
essential.

\begin{table}[H]
    \centering
    \caption{\textbf{Comparison with explicit methods under monocular
    setting (CD $\downarrow$).} Explicit methods diverge even with
    depth supervision; \OM remains robust.}
    \label{tab:explicit_comparison}
    \begin{tabular}{llccc}
        \toprule
        Method & Supervision & Ball & Cat & Status \\
        \midrule
        PAC-NeRF~\cite{li2023pac} & Multi-view & 0.85 & 0.29
        & Reference \\
        NeuMA~\cite{neuma} & Multi-view & 0.98 & 0.35
        & Reference \\
        \midrule
        PAC-NeRF + Depth & Monocular & $>$100 & $>$100
        & Diverged \\
        NeuMA~\cite{neuma} & Monocular & 1.12 & 0.84
        & Suboptimal \\
        \textbf{Ours} & Monocular & \textbf{0.92} & \textbf{0.32}
        & Robust \\
        \bottomrule
    \end{tabular}
\end{table}

\begin{figure}[H]
    \centering
    \begin{minipage}[c]{0.62\linewidth}
        \centering
        \includegraphics[width=\linewidth]{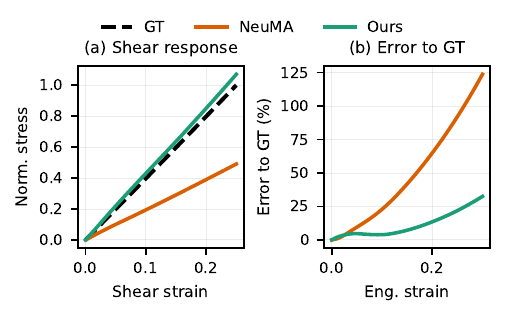}
    \end{minipage}
    \hfill
    \begin{minipage}[c]{0.34\linewidth}
        \centering
        \small
        \setlength{\tabcolsep}{3pt}
        \renewcommand{\arraystretch}{0.92}
        \begin{tabular}{lccc}
            \toprule
            Method & Uni. & Sh. & Avg. \\
            \midrule
            NeuMA~\cite{neuma} & 52.0 & 27.1 & 39.6 \\
            \textbf{Ours} & \textbf{12.0} & \textbf{3.9} & \textbf{8.0} \\
            \bottomrule
        \end{tabular}

        \vspace{2pt}
        {\footnotesize Response error (\%) $\downarrow$}
    \end{minipage}
    \caption{\textbf{Direct law-level validation on held-out deformation gradients.}
    Left: normalized constitutive response under analytic JellyDuck ground truth.
    Right: quantitative response error. \OM more accurately recovers the
    underlying constitutive response than NeuMA, reducing the average error
    from 39.6\% to 8.0\%.}
    \label{fig:lawlevel}
\end{figure}

\noindent\textbf{Direct law-level validation.}
Beyond trajectory-level alignment, we further evaluate whether the learned
implicit constitutive law matches the underlying physical response. Using an
analytic JellyDuck law as ground truth, we query the learned elasticity under
held-out deformation gradients that are not used for image supervision.
As shown in~\cref{fig:lawlevel}, \OM follows the ground-truth response more
closely than NeuMA and reduces the average constitutive response error from
39.6\% to 8.0\%. This indicates that the improvement is not merely due to
better trajectory fitting, but also to more accurate law-level recovery.

\noindent\textbf{Robustness to out-of-distribution materials.}
A critical concern is whether CPR becomes counterproductive when the
true material is absent from the hypotheses.
\Cref{tab:ood} addresses this with two stress tests: (1)~a
\textit{blind test} where the ground-truth constitutive model is
deliberately removed from the library, and (2)~\textit{composite
materials} combining elasticity, plasticity, and fluid
properties---behaviors not covered by any single hypothesis. In both
cases, \OM still substantially outperforms the baseline, confirming
that CPR provides \emph{general physical guidance} (\eg, energy
consistency, stability constraints) rather than requiring exact
template matches. The implicit LoRA layers successfully compensate for
residual differences between the prior and actual material behavior.

\begin{table}[H]
    \centering
    \caption{\textbf{Robustness to out-of-distribution materials
    (CD $\downarrow$).} \OM outperforms baselines even when the
    correct hypothesis is absent or for composite materials.}
    \label{tab:ood}
    \begin{tabular}{lccc}
        \toprule
        Configuration & Elastomer & Plasticine & Composite \\
        & (target excl.) & (target excl.) & (undefined) \\
        \midrule
        NeuMA~\cite{neuma} & 1.123 & 0.844 & 3.628 \\
        Ours (Target Removed) & 0.976 & 0.397 & -- \\
        \textbf{Ours (Full Library)} & \textbf{0.922} &
        \textbf{0.318} & \textbf{1.275} \\
        \bottomrule
    \end{tabular}
\end{table}

\noindent\textbf{Generalization.}
The learned constitutive laws are not tied to the specific geometry used
during training: once acquired, they can be applied to novel object shapes
and further support multi-object interaction scenarios where different
objects follow different learned material laws. Additional qualitative
results are provided in the supplementary material.

\section{Conclusion}

We presented \OM, a framework for learning implicit constitutive laws from monocular dynamic video through visual-physical bidirectional alignment. By unifying LoRA-based adaptation with Rank-based Depth-Geometric Anchors (RDGA) and a Constitutive Prior Regularizer (CPR), our method achieves 48\% lower Chamfer Distance than the strongest baseline on synthetic data, strong generalization on real-to-sim datasets, and superior quality in real-world monocular experiments---while remaining robust even when the true material is absent from the hypotheses.

\noindent\textbf{Limitations.}
GCA still requires a static multi-view scan for geometric initialization, and extending the entire pipeline to fully monocular reconstruction is left for future work. Its geometric anchors also depend on the quality
of monocular depth estimation, although the rank-based formulation mitigates scale-shift ambiguity. Finally, our current experiments focus mainly on single-object dynamics; extending the framework to dense multi-object scenes with heterogeneous materials is a promising
direction.

\bigbreak
\noindent\textbf{Acknowledgements.} This work is supported by Hong Kong Research Grants Council -- General Research Fund (Grant No.\ 17213825), Hong Kong Innovation and Technology Commission -- Innovation and Technology Fund (Grant No.\ ITS/488/24FP), and HKU Seed Fund for PI Research.

\bibliographystyle{splncs04}
\bibliography{main}

\end{document}